\documentclass[10pt,twocolumn,letterpaper]{article}
\usepackage{verbatim}
\usepackage{wacv}
\usepackage{times}
\usepackage{epsfig}
\usepackage{graphicx}
\usepackage{amsmath}
\usepackage{amssymb}
\usepackage{mathtools}
\usepackage{authblk}

\makeatletter
\newcommand{\printfnsymbol}[1]{%
  \textsuperscript{\@fnsymbol{#1}}%
}
\makeatother

\def\wacvPaperID{1277} 

\wacvfinalcopy 
\ifwacvfinal
\def\assignedStartPage{9876} 
\fi

\ifwacvfinal
\usepackage[breaklinks=true,bookmarks=false]{hyperref}
\else
\usepackage[pagebackref=true,breaklinks=true,colorlinks,bookmarks=false]{hyperref}
\fi

\ifwacvfinal
\else
\fi

\begin{document}

\title{A Variational Information Bottleneck Based Method to Compress Sequential Networks for Human Action Recognition}

\author{\textbf{Ayush Srivastava}\thanks{equal contribution}\textsuperscript{\small{\: 1}}
\quad
\textbf{Oshin Dutta}\printfnsymbol{1}\textsuperscript{\small{1}}
\quad
\textbf{Jigyasa Gupta}\textsuperscript{\small{2}}
\quad
\textbf{Sumeet Agarwal}\textsuperscript{\small{1}}
\quad
\textbf{Prathosh AP}\textsuperscript{\small{1}}\\
\textsuperscript{\small{1}}Indian Institute of Technology Delhi\\
\textsuperscript{\small{2}}Samsung R\&D Institute India - Delhi\\
{\tt\small ayush.srivastava@ee.iitd.ac.in, oshin.dutta@ee.iitd.ac.in, jigyasa.g@samsung.com, sumeet@ee.iitd.ac.in, prathoshap@ee.iitd.ac.in}
}


\maketitle
\begin{abstract}
   In the last few years, deep neural networks' compression has become an important strand of machine learning and computer vision research. Deep models require sizeable computational complexity and storage when used, for instance, for Human Action Recognition (HAR) from videos, making them unsuitable to be deployed on edge devices. In this paper, we address this issue and propose a method to effectively compress Recurrent Neural Networks (RNNs) such as Gated Recurrent Units (GRUs) and Long-Short-Term-Memory Units (LSTMs) that are used for HAR. We use a Variational Information Bottleneck (VIB) theory-based pruning approach to limit the information flow through the sequential cells of RNNs to a small subset. Further, we combine our pruning method with a specific group-lasso regularization technique that significantly improves compression. The proposed techniques reduce model parameters and memory footprint from latent representations, with little or no reduction in the validation accuracy while increasing the inference speed several-fold. We perform experiments on the three widely used Action Recognition datasets, viz. UCF11, HMDB51, and UCF101, to validate our approach. We show that our method achieves over 70 times greater compression than the nearest competitor with comparable accuracy for action recognition on UCF11.
\end{abstract}

\section{Introduction}
Recent years have witnessed tremendous progress in embedded and mobile devices, such as unmanned drones, automatic cars, smart devices, and smart glasses. The demand to deploy highly accurate Deep Neural Network (DNN) models on these devices has become much more intense than ever. However, some essential resources in these devices, such as the storage,  computational units, and battery power, are limited, which pose several challenges in incorporating large DNNs under low-cost/resource settings.

The DNNs specifically used for HAR are usually much more resource-intensive than those used for other applications since HAR requires both spatial and temporal information processing. Many DNN architectures have been proposed for HAR, Convolutional Neural Network- LSTM (CNN-LSTM) \cite{shi2015convolutional}, being one of the major ones. It comprises a Deep Convolutional Neural Network (CNN) feature extractor, followed by an RNN, specifically an LSTM unit.
A significant amount of research work focuses on CNN architectures' compression, but RNN architectures' compression is still an active research area.
RNN model compression is beneficial for efficient inference because sequential processing of data through the time-steps takes significant time during inference, especially in CNN-RNN architectures. Further, many sequential models are largely over-parameterized, and a large number of parameters often lead to overfitting. Addressing these issues in this paper, we focus on the compression of LSTM for CNN-LSTMs and end-to-end LSTM architectures, but the method can be easily generalized to other RNN architectures as well.

Most of the research work that has been done on RNN compression and acceleration involves either (i) matrix factorization or tensor decomposition \cite{tensorR, yang2017tensor, yin2020compressing, ye2018learning}, or (ii) an unconventional deep learning architecture \cite{diba2017temporal, zhu2018end}. Other RNN compression approaches include regularization based method \cite{wen2018learning}, knowledge distillation \cite{shi2019knowledge} and parameter quantization \cite{alom2018effective}. Methods based on tensor decomposition capitalizes on the fact that the large size of an RNN is primarily due to the input-to-hidden matrix of the RNN due to the large input size. This is especially seen in end-to-end RNN training where the input stream is the RGB values of images, and hence, they focus on the compression of parameters in the input-to-hidden layers.
A group lasso regularization based LSTM compression method \cite{wen2018learning} implements group weight regularization and prunes hidden states through the intrinsic sparsity in the LSTM structure. 
It aims to reduce only the hidden state vector's size, while the input vector to LSTM is usually much larger than the hidden state vector. In contrast, we aim to reduce both the input dimension and the number of hidden states of LSTMs.

Specifically, we propose a Variational Information Bottleneck (VIB) based method to compress RNNs. The idea of information bottlenecking introduced by \cite{tishbyIB} and its variational form \cite{alemi} has been successfully utilized to remove neurons in CNN and linear architectures \cite{daiIB}.  Our approach proposes a novel way to reduce input feature dimensions and hidden states in RNNs by learning compressed latent representations.
Our method does not require additional VIB parameters during inference, just the usual LSTM parameters. It requires minimal hyperparameter tuning and focuses on reducing the complete LSTM structure rather than just the input-to-hidden matrix as done in previous work. Furthermore, none of the previous LSTM compression techniques have explicitly shown the results of combining multiple compression methods. We evaluate end-to-end VIB-LSTM architecture and CNN based VIB-LSTM architecture on the three most popular action recognition datasets. The proposed approach achieves the most compression for better or comparable accuracy. The significant contributions of this paper are the following:
\begin{enumerate}
    \item We propose a novel VIB-LSTM structure that trains high accuracy sparse LSTM models.
    \item We develop a principled sequential network compression pipeline that sparsifies pre-trained model matrices of RNNs/LSTMs/GRUs.
    \item We develop a new VIB framework to compress CNN-LSTM based architectures specifically.
    \item We evaluate our method on popular action recognition datasets to yield compact models with validation accuracy comparable to that of state-of-the-art models.
\end{enumerate}
\section{Related Work}
\subsection{Human Action Recognition}
Approaches to human action recognition (HAR) involve either using handcrafted features and machine learning or using end to end deep learning. Spatio Temporal Interest Point(STIP) methods \cite{nazir2018evaluating} of HAR extend the idea of object feature detection to the 3D domain. Although these methods are rotation, scale, illumination invariant, they lead to erroneous classification with camera movements. Trajectory based HAR \cite{gaidon2014activity,wang2016robust} track specific vital points in the video to classify actions. These methods adapt to changing camera angles but require 2-D or 3-D skeletal joint points.  
Deep CNN based two-stream architectures take RGB data and optical flow, calculated from the images, as inputs. Motion vectors \cite{zhang2018real} instead of optical flows, change of fusion point \cite{feichtenhofer2016convolutional}, and Temporal Segment Networks improve two-stream methods of HAR. Calculation of flow requires additional processing power. Inflating 2-D CNNs lead to 3-D CNNs \cite{carreira2017quo}, 3D ConvNet \cite{diba2017temporal}, and 3-D two stream \cite{zhu2018end} which yield high accuracy of action classification, at the cost of tens of millions of parameters \cite{carreira2017quo}, and hundreds of GFLOPs; not suited to applications on edge.

CNNs pipelined with LSTMs capture long term temporal dependencies of CNN extracted features as done by methods combining CNNs with stacked LSTMs \cite{yue2015beyond} and Long term recurrent convolutional networks(LRCN) \cite{donahue2015long}. 2-D Conv-LSTM architectures have lower parameters, low memory footprint from hidden representations, and higher inference speed than 3-D networks with comparable accuracy. Over-parameterized LSTMs in such architectures lead to overfitting and large resource requirements. Our compressed CNN-VIB-LSTM models need lower parameters with comparable accuracy to the 3-D counterparts.
\subsection{LSTM compression}
Approaches to reducing the size of RNN/LSTM/GRU parameters include low-rank matrix factorization and hybrid matrix factorization with improved accuracy over Low-Rank Matrix Factorization. An improvement over factorization methods, the tensor decomposition methods \cite{tensorR}\cite{yang2017tensor} reshape input vector and weight matrices into tensors. As a contrast, our method reduces the size of input features by retaining features only relevant to prediction. Compression of hidden states and gate outputs in \cite{wen2018learning} yield sparse structures in LSTM through group lasso regularization. Unlike our method, it does not consider compression of high dimensional input induced large input-to-hidden transformation matrix of the LSTMs.   

Overall parameter reduction of LSTMs has been attempted by \cite{dai2019grow}, who add in a linear layer with activation after LSTM gates, which controls the joint LSTM weight matrix's sparsification through magnitude pruning. Our method differs from their work since our pruning method works with an information-theoretic perspective rather than the magnitude of weights. Moreover, unlike their model, our compression inducing layers added during training are removed in the compressed smaller model to be used for inference while still maintaining accuracy. \cite{daiIB} proposes VIB based method to prune out neurons from fully trained models yielding sparse structured fully connected and convolution layers. 

\section{The Proposed Approach}

\subsection{Notations and Preliminaries}

\begin{itemize}
    \item $\mathbf{X} \in {\mathbb{R}^{m_1 \times m_2 \times 3 \times T}}$: Input tensor consisting of $T$ stacked RGB frames (each RGB frame of dimension $m_1 \times m_2 \times 3$) from a video.
    \item $\mathbf{Y} \in \{1,2,...,a\}$: Output action class vector, with $a$ being number of actions.
    \item $\mathcal{D}$: Joint distribution between $\mathbf{X}$ and $\mathbf{Y}$. 
    \item $\mathbf{v} \in {\mathbb{R}^{d \times T}}$: The $d$-dimensional output of the Convolutional feature extractor block of CNN-LSTM, corresponding to $T$ frames, with $\mathbf{v}^{t} \in {\mathbb{R}^{d}}$ denoting its $t^{th}$ frame. Note that $\mathbf{v}^{t}$ is the input to the LSTM.
    \item$\mathbf{h}^{t} \in {\mathbb{R}^{n}}$ : Hidden state vector or output of the LSTM corresponding to the $t^{th}$ frame. Also, $\mathbf{h}^{T}$ denotes hidden state vector corresponding to the last time-step. 
    \item $\mathbf{k}^{T} \in \{\mathbf{i}^{T}, \mathbf{f}^{T}, \mathbf{o}^{T}, \mathbf{g}^{T}\}$: Usual notations for LSTM gate outputs corresponding to the last time step.
\end{itemize}

\subsection{Variational Information Bottle-necking (VIB) }



Our goal is to learn a compressed representation $\tilde{\mathbf{k}}^{T} \in \mathbb{R}^{l}$ of  $\mathbf{k}^{T} \in \mathbb{R}^{n}$ ($l \ll n$) $ \forall \mathbf{k}^{T}$ while retaining relevant spatial and temporal information in $\mathbf{v}$ required for prediction. Note that compressing $\mathbf{k}^{T}$ is equivalent to compressing $\mathbf{h}^{T}$ since there is a deterministic mapping between them. In Variational Information Bottle-necking (VIB) framework \cite{alemi}, this is cast as an optimizatiom problem where the goal is to learn $\tilde{\mathbf{k}}^{T}$ such that it has least information regarding the LSTM input $\mathbf{v}$ while retaining all the relevant information needed for learning the target $\textbf{Y}$. Mathematically, it amounts to optimizing the following objective function:
\begin{equation}
\label{e1}
     \mathcal{L} = \min _{\theta} \sum_{\tilde{\mathbf{k}}^{T}  \in \{ \mathbf{i}^{T}, \mathbf{f}^{T}, \mathbf{o}^{T}, \mathbf{g}^{T} \}} \beta I(\tilde{\mathbf{k}}^{T},\mathbf{v} ; \theta)- I(\tilde{\mathbf{k}}^{T},\mathbf{Y} ; \theta) 
\end{equation} 
where $I(\cdot)$ denotes mutual information between two random variables and $\theta$ is the parameter set of a compression neural network that transforms  $\mathbf{v}$ to $\tilde{\mathbf{k}}^{T}$, $\beta$ is a hyper-parameter which controls the amount of trade-off between compression and prediction accuracy. To avoid notational complexity, we denote compressed version $\tilde{\mathbf{k}}^{T}$ by ${\mathbf{k}^{T}}$ hereafter.
Eq. \ref{e1} is intractable in general because of the model complexity and infeasibility of the mutual information term. Thus, a variational upper bound, ${\tilde{\mathcal{L}}}$ \cite{alemi} is invoked on it:
\begin{equation}
\label{e2}
\begin{split}
    \tilde{\mathcal{L}} = \sum_{\mathbf{k}^{T}} \mathbb{E}_{\mathbf{X}, \mathbf{Y},\mathbf{v},\mathbf{h}^T} \bigg[\beta \mathbb{D}_{KL}\Big[p\big({\mathbf{k}^{T}} \mid \mathbf{v}\big) \| q\big({\mathbf{k}^{T}} \big)\Big]\\ \vspace{0.3 in}
    - \log q\big({\mathbf{Y} \mid {\mathbf{h}^T}}\big)\bigg] \geq {\mathcal{L}}
\end{split}
\end{equation} 
where $q(\mathbf{k}^T)$ and $q(\mathbf{Y} \mid \mathbf{h}^T)$ denote two variational distributions which respectively approximate $p(\mathbf{k}^T \mid \mathbf{v})$ and $p(\mathbf{Y} \mid \mathbf{h}^T)$ (Refer to Appendix for the derivation). The first term in Eq. \ref{e2} approximates the amount of relevant information captured from $\mathbf{v}$ while the second term is responsible for maintaining the prediction accuracy.



\subsection{Compression using VIB}

With the aforementioned framework, the compressed $\mathbf{k}^{T}$ is defined as follows:
\begin{equation}
\label{e3}
\begin{array}{cc}
    \mathbf{k}^{T} =  \mathbf{z}_{k} \odot f_{k} (\mathbf{v}) & ; 
    \quad \mathbf{z}_k = \mathbf{\mu}_{k} + \mathbf{\epsilon} \odot \mathbf{\sigma}_{k} 
\end{array}
\end{equation} 
where $\odot$ denotes element-wise multiplication, ${k} \in \{ \mathbf{i}, \mathbf{f}, \mathbf{o}, \mathbf{g} \}$, $\mathbf{z}_{k}$ is a random shared vector,  $\mathbf{\mu}_{k}$ and $\mathbf{\sigma}_{k}$ are learnable parameters which are shared across time while $\mathbf{\epsilon}$ is sampled from an isotropic  Gaussian distribution and $f(\cdot)$ denoting the standard LSTM equations. With these definitions and assuming the elements of $\mathbf{k}^{T}$ to be uncorrelated, $p({\mathbf{k}}^{T} \mid \mathbf{v}^{t} )$ takes the following form:
\begin{equation}
\label{e4}
     p({\mathbf{k}}^{T} \mid \mathbf{v}  )=\mathcal{N}\bigg({\mathbf{k}}^{T} ;    f_{k}(\mathbf{v}) \odot \mathbf{\mu}_{k},\\ \text{diag}[f_{k}(\mathbf{v})^{2} \odot \mathbf{\sigma}_{k}^{2}]\bigg)
 \end{equation}
Further, we assume $q(\mathbf{k}^T)$ also as an Isotropic Gaussian distribution with zero mean and learnable vector of variances, $\mathbf{\xi}_\mathbf{k}$ \cite{daiIB}.
\begin{align}
\label{e5}
    q(\mathbf{k}^T) = \mathcal{N} \Big(\mathbf{k}^T; 0,\text{diag}\big[\mathbf{\xi}_\mathbf{k}\big]\Big)
\end{align}
Note that if any $j^{th}$ element $\mathbf{\xi}_{\mathbf{k}j}$ of the variance vector $\mathbf{\xi}_\mathbf{k}$ is pushed to zero during training, corresponding dimension of $p({\mathbf{k}}^{T} \mid \mathbf{v})$ is pushed to be a degenerate dirac-delta, making that dimension redundant. Assuming that $\mathbf{\xi}_\mathbf{k}$ can be learnt optimally, KL term in the Eq. \ref{e2} takes a closed form expression, simplifying (Refer to Appendix) the loss in Eq. \ref{e2} as follows:
\begin{equation}
\label{e6}
\begin{array}{c}
\begin{aligned}
    \tilde{\mathcal{L}}=\sum_{\mathbf{k}}\beta \sum_{j=1}^{l} \Big[\log (1+ \frac{\mathbf{\mu}^2_{\mathbf{k}j}}{\mathbf{\sigma}^2_{\mathbf{k}j}}) + \psi_{\mathbf{k}j} \Big]\\
    - 4\mathbb{E}_{\mathbf{X}, \mathbf{Y},\mathbf{v},\mathbf{h}^T}\Big[\log q({\mathbf{Y} \mid {\mathbf{h}^T}})\Big]
\end{aligned}
\end{array}
\end{equation}
where, $\psi_{\mathbf{k}j} \geq 0$ by Jensen's Inequality and is given by:
\begin{equation}
\label{e0}
    \psi_{\mathbf{k}j} = \log \Big(\mathbb{E}_{\mathbf{X}, \mathbf{Y},\mathbf{v}}\Big[f_{\mathbf{k}j}(\mathbf{v})^2\Big]\Big)
    - \mathbb{E}_{\mathbf{X}, \mathbf{Y},\mathbf{v}}\Big[\log\Big(f_{\mathbf{k}j}(\mathbf{v})^2\Big)\Big]
\end{equation}
Note from Eq. \ref{e6} that our loss function for LSTM is similar to that obtained for CNN \cite{daiIB}. However, \cite{daiIB} used it for dimensionality reduction of linear and CNN layers while ours is specifically for LSTM compression. Also, $\psi_{\mathbf{k}j}$ can be assumed to be sufficiently small as removing it from  Eq. \ref{e6} does not affect our results.  Although the expectation in the second term of the Eq. \ref{e6} does not have a closed-form, its unbiased stochastic approximation such as a monte-carlo estimate can be used for training the network \cite{Kingma2014AutoEncodingVB}.




\subsection{VIB-LSTM - Implementation}

\begin{figure}
\begin{center}
\includegraphics[width=8cm,scale=1.5]{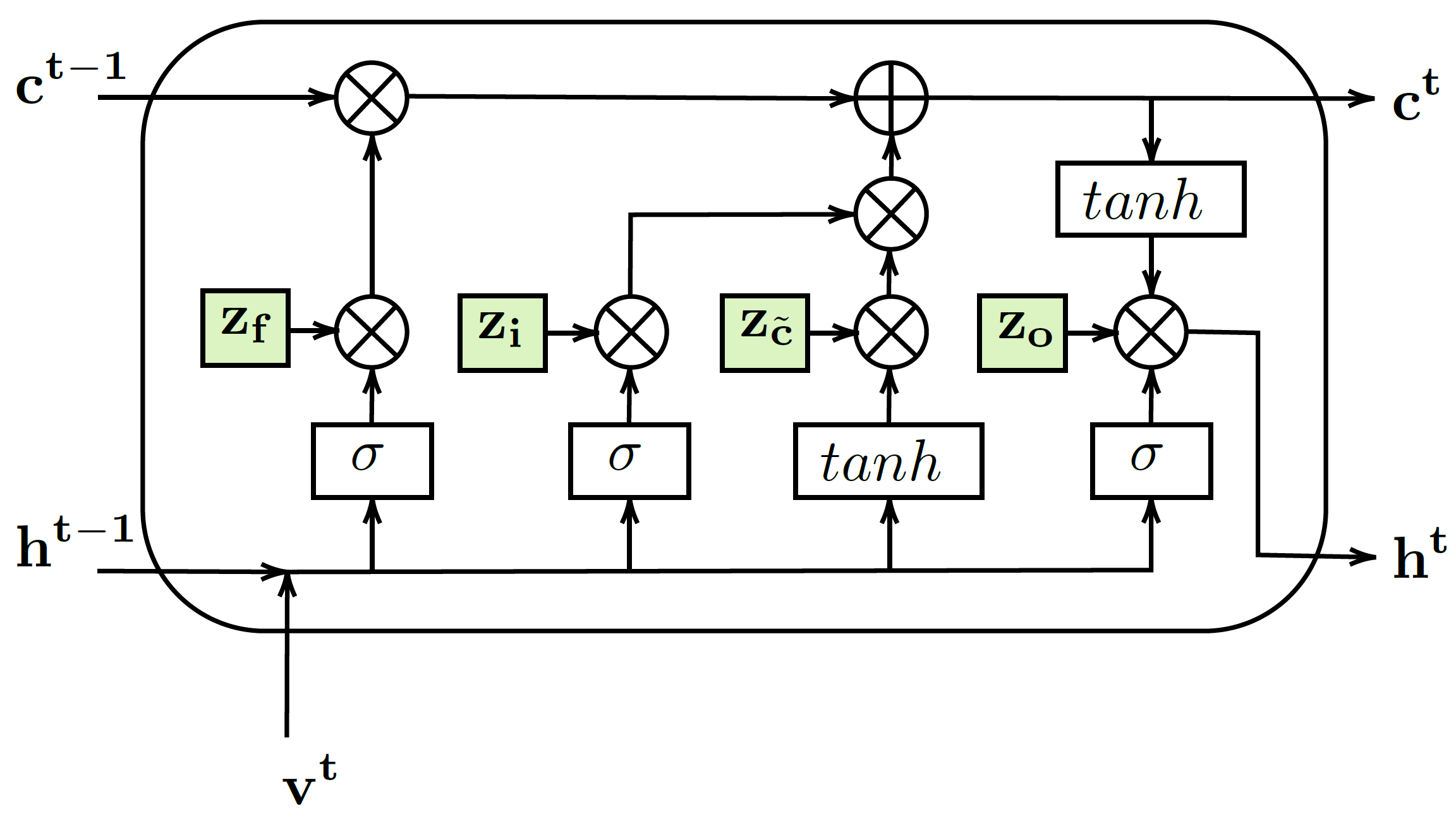}
\end{center}
   \caption{A single VIB-LSTM cell. It has a VIB layer $\mathbf{z}_k$ (shown in green) after every gate output, which acts as a trainable mask. Each $\mathbf{z}_k$ follows a multivariate Gaussian distribution with trainable parameters. }
\label{fig:lstmvib}
\end{figure}

We propose to apply the previously mentioned VIB principle (Eq. 3) as a layer to the LSTM gate outputs which gives rise to the following equations for VIB-LSTM:
\begin{equation}
\label{e7}
\begin{array}{l}
\mathbf{i}^{t}=\mathbf{z}_i \odot \sigma(\mathbf{W}_{ix}\cdot \mathbf{v}^{t}+\mathbf{U}_{i h} \cdot \mathbf{h}^{t-1}+\mathbf{b}_{i}) \\
\mathbf{f}^{t}=\mathbf{z}_f \odot \sigma(\mathbf{W}_{fx}\cdot \mathbf{v}^{t}+\mathbf{U}_{fh} \cdot \mathbf{h}^{t-1}+\mathbf{b}_{f}) \\
\mathbf{o}^{t}=\mathbf{z}_o \odot \sigma(\mathbf{W}_{ox}\cdot \mathbf{v}^{t}+\mathbf{U}_{oh} \cdot \mathbf{h}^{t-1}+\mathbf{b}_{o}) \\
\mathbf{g}^{t}={\mathbf{z}_{g}} \odot \tanh(\mathbf{W}_{gx}\cdot \mathbf{v}^{t}+\mathbf{U}_{gh} \cdot \mathbf{h}^{t-1}+\mathbf{b}_{g}) \\
\mathbf{\tilde{c}}^{t}=\mathbf{f}^{t} \odot \mathbf{\tilde{c}}^{t-1}+\mathbf{i}^{t} \odot \mathbf{g}^{t} \\
\mathbf{h}^{t}=\mathbf{o}^{t} \odot \tanh (\mathbf{\tilde{c}}^{t})
\end{array}
\end{equation}
where $\mathbf{z}_i$, $\mathbf{z}_f$, $\mathbf{z}_o$ and ${\mathbf{z}_g}$ are the trainable VIB masks shared across time, for each of the LSTM gates as shown in Figure \ref{fig:lstmvib}. Note that the structure of VIB-LSTM is very similar to the standard LSTM structure, which makes VIB-LSTM easy to train.


\subsection{LSTM pruning with VIB mask}

The VIB masks mentioned in the previous section can be learned for each gate to push down the redundant elements of the gate outputs close to zero, which in turn will push the corresponding elements of the hidden state vector close to zero. We propose a simple threshold on the following ratio to prune the parameters of four LSTM gates, using the :
\begin{equation}
\label{e8}
    \alpha_{\mathbf{k}j} = \mu_{\mathbf{k}j}^2 \sigma_{\mathbf{k}j}^{-2}
\end{equation} 
The rationale for the aforementioned threshold comes from the following proposition (\cite{daiIB}):

\textbf{Proposition-1}: At any minimum of Eq. \ref{e6}, $\alpha_{\mathbf{k}j} = 0$ is a necessary condition for $I(\mathbf{k}^T_j, \mathbf{v}) = 0$ and a sufficient condition for $I(\mathbf{k}^T_j, \mathbf{v}) \leq \psi_{\mathbf{k}j}$. (Proof given in \cite{daiIB})

The above proposition implies that for minimizing the mutual information and ignoring the $\psi$ term in Eq. \ref{e6}, it is necessary to set $\alpha_{\mathbf{k}j}$ to small values. 
Therefore, in summary, the units $\mathbf{k}^T_j$ of $\mathbf{k}^T$ for which $\alpha_{\mathbf{k}j} = 0$, do not contribute significantly in propagating relevant information for prediction and thus, are valid candidates to be pruned. Consequently, the LSTM hidden state units, LSTM cell state units and the LSTM parameters connected to such neurons also become redundant. Therefore, LSTM parameter matrix can be reduced in size by removing the corresponding redundant rows and columns of the matrix (Refer to Appendix for more details). Finally, the trainable masks are not required to be stored, and a standard LSTM with reduced hidden vector size can be used for inference.


\subsection{VIB pruning for Convolutional Features}

To ensure further compression, we propose to use a similar VIB principles on the output of the convolutional features $\mathbf{v}$, that forms the input to the LSTM layers. This can be cast as the following objectively similar to Eq. \ref{e1}:
\begin{equation}
\label{e9}
    \mathcal{L}_v = \min _{\theta_v} \beta_v I(\mathbf{v},\mathbf{x} ; \theta_v)- I(\mathbf{v},\mathbf{Y} ; \theta_v) 
\end{equation} Again we invoke a variational upper bound on Eq. \ref{e9} (refer to the Appendix for the derivation):
\begin{equation}
\label{e10}
\begin{split}
    \tilde{\mathcal{L}_v} = \mathbb{E}_{\mathbf{X}, \mathbf{Y},\mathbf{v},\mathbf{h}^T}\bigg[ \beta_v \mathbb{D}_{KL}\Big[p\big(\mathbf{v} \mid \mathbf{x}\big) \| q\big({\mathbf{v}} \big)\Big]\\ \vspace{0.3 in}
    - \log q\big({\mathbf{Y} \mid {\mathbf{h}^T}}\big)\bigg] \geq {\mathcal{L}_v}
\end{split}
\end{equation} 
The KL-divergence term in Eq. \ref{e10} can be simplified after re-parameterizing $\mathbf{v}$ using learnable parameters ($\mu_v, \sigma_v$) and by assuming gaussian distributional forms for $p(\mathbf{v} \mid \mathbf{X})$ and $q(\mathbf{v})$, which reduces the loss function as follows:
\begin{equation}
\label{e11}
p(\mathbf{v} \mid \mathbf{X}) =\mathcal{N}\bigg(\mathbf{v};    f_v(\mathbf{X}) \odot \mathbf{\mu}_v, \text{diag}[f_v(\mathbf{X})^{2} \odot \mathbf{\sigma}_v^{2}]\bigg) \nonumber
\end{equation}
\begin{equation}
    q(\mathbf{v}) = \mathcal{N} \Big(\mathbf{v}; 0,\text{diag}\big[\mathbf{\xi}_v\big]\Big)
\end{equation}
\begin{equation}
    \tilde{\mathcal{L}_v}=\beta_v \sum_{j}\log (1+ \frac{\mathbf{\mu}^2_{vj}}{\mathbf{\sigma}^2_{vj}}) - \mathbb{E}_{\mathbf{X}, \mathbf{Y},\mathbf{v},\mathbf{h}^T}\Big[\log q({\mathbf{Y} \mid {\mathbf{h}^T}})\Big]
\end{equation}
Following Proposition-1 with similar arguments as with VIB-LSTM, redundant units $\mathbf{v}_j$ of output feature vector or the VIB-LSTM input $\mathbf{v}$ can be pruned by using some threshold on the following quantity:
\begin{equation}
\label{e12}
    \alpha_{vj} = \mu_{vj}^2 \sigma_{fv}^{-2}
\end{equation} 
Finally, the overall loss function for our method is obtained by by adding Eq. \ref{e6} and Eq. \ref{e11} as follows:
\begin{equation}
\begin{split}
\label{e14}
     \mathcal{L}_{total} = \sum_{\mathbf{k}}\beta \sum_{j=1}^{l}\log (1+ & \frac{\mathbf{\mu}^2_{\mathbf{k}j}}{\mathbf{\sigma}^2_{\mathbf{k}j}}) 
     +  \beta_v \sum_{j}\log (1+ \frac{\mathbf{\mu}^2_{vj}}{\mathbf{\sigma}^2_{vj}}) \\
     &- 5\mathbb{E}_{\mathbf{X}, \mathbf{Y},\mathbf{v},\mathbf{h}^T}\Big[\log q({\mathbf{Y} \mid {\mathbf{h}^T}})\Big]
\end{split}   
\end{equation}
Eq. \ref{e14} shows that our overall loss function is simply the cross-entropy loss with regularization terms constraining the information flow to subsets of units of specific intermediate layers.
Therefore, we use a VIB mask on feature extractor output along with VIB-LSTM, and we aim to reduce the dimensionalities of both $\mathbf{v}^t$ and $\mathbf{h}^t$, which may result in a massive reduction of LSTM parameters.

\subsection{Group Lasso Regularization}
Finally, in addition to the proposed method, we also apply an existing hidden state pruning method \cite{wen2018learning} where a structured group lasso regularization is used to push down the LSTM parameter groups to zero. This method utilizes the intrinsic sparse structure (ISS) of the LSTM for pruning redundant hidden state. The corresponding regularization term takes the form: 
\begin{equation}
\label{e15}
\begin{array}{c}
\begin{aligned}
R(\mathbf{W})=\sum_{k}^{K}\|w_{k}\|_{2}
\end{aligned}
\end{array}
\end{equation}
where $\mathbf{W}$ is the tensor containing all the LSTM parameters, n is the total number of weight groups equal to the size of the hidden state vector and $\|.\|_{2}$ is the \textit{l2}-norm. The weight groups are pruned based on a threshold.

\section{Datasets and Models}
\subsection{Datasets}
We have evaluated our proposed method of LSTM compression on three popular human action recognition datasets- UCF11 \cite{liu2009UCF11}, HMDB51 \cite{kuehne2011hmdb} and UCF101 \cite{soomro2012ucf101}. UCF11 comprises videos from youtube and  represents 11 classes of actions: basketball shooting, biking/cycling,diving, golf swinging among others. 
HMBD51 contains 6849 clips classified into 51 action classes, taken from movies and other public databases. The dataset having large variations within classes contains daily life activities such as brushing hair,  sitting, and standing besides sports clips. 
UCF101 is one of the most challenging datasets due to large variations in camera motion, object appearance and pose, object scale, viewpoint, cluttered background, and illumination conditions. The data consists of 13320 videos of 101 action classes, including Cricket Bowling, Cricket Shot, Playing Sitar, Pole Vault, Sumo Wrestling, etc.

\subsection{Baseline models}
For end-to-end LSTM experiments, we compare our method to tensor decomposition based LSTM compression techniques \cite{tensorR, yin2020compressing, ye2018learning, yang2017tensor}. For our experiments on LSTM compression for CNN-LSTM based architectures, we compare our approach with baseline (uncompressed model), namely Naive-LSTM, and with previous LSTM compression methods, namely tensor ring decomposition method \cite{tensorR} and a group lasso regularization based compression technique, namely ISS \cite{wen2018learning}.
We also compare with formerly state-of-the-art architectures used for action recognition. Two-stream LSTM network architecture \cite{twolstm} uses both RGB and optical flows as input to the LSTM. Attention-based ConvLSTM \cite{ge2019attention} proposes attention based improvement over CNN-LSTM. Two-stream I3D \cite{carreira2017quo} introduces inflated 3-D convolutions to implicitly model the temporal information in videos.

\subsection{Implementation details}
\textbf{End-to-end LSTM Training}: For our experiments with end-to-end LSTM, we use the same configurations as in \cite{tensorR}. We sample 6 RGB frames from each video and use $160 \times 120 \times 3 = 57600$ dimensional vector, derived from each frame, as input to the LSTM. The LSTM has a 256-dimensional hidden state vector and is followed by a Fully-Connected (FC), Batch Normalization (BN), and softmax layer.

\textbf{LSTM training with pre-trained CNN}: We take imagenet-pre-trained convolutional networks to extract features from the videos for our LSTM compression experiments with CNN-LSTM architectures. For training our CNN-VIB-LSTM models, we use two different learning rates: one for VIB parameters varying from $1 \times 10^{-1}$ to $1\times10^{-3}$, and the other varying from $\times 10^{-3}$ to $\times 10^{-5}$ for other model parameters. We use dropout with probability 0.5 to avoid overfitting in most of our experiments.

\textbf{Training with UCF11}: For training our models with UCF11, we take a randomly sampled sequence of 32 frames per video, i.e., 1.06 seconds from each video. Instead of 32 or more frames, as done in the previous approaches \cite{tensorR}, we note that training with eight frames per video does not lead to any accuracy degradation and makes the training faster. Image size is kept at 299$\times$299 for input to imagenet pretrained-Inception\_v3 \cite{szegedy2016incepv3} and $240\times240$ for input to imagenet pre-trained- Efficientnet-b1 \cite{tan2019efficientnet}. 

\textbf{Training with HMDB51 and UCF101}: For training our models with HMDB51 and UCF101, we use both RGB and TV-L1  optical flow \cite{zach2007duality}. We randomly crop the video frames and resize them to $240\times240$. For both HMDB51 and UCF101, we fuse the RGB and optical flow features and pass as input to the LSTM. We re-scale the values of the optical flow to [0,255]. All the images are center cropped and normalized for batching of the data.


\begin{table}[t]
\begin{center}
\begin{tabular}{p{2.2cm}|p{1.5cm}|p{1.7cm}|p{1cm}}
\hline
\textbf{\small Methods}&\textbf{\small LSTM}&\textbf{\small LSTM}&\textbf{\small Accuracy}\\
&\textbf{\small Parameters}&\textbf{\small Compression Ratio}\\
\cline{1-4}
Naive-LSTM & 59.246M & 1 &69.7$\%$  \\
\small TT-LSTM \cite{yang2017tensor}	&0.268M & 221.0 & 79.6$\%$ \\
\small BT-LSTM \cite{ye2018learning} &	0.268M & 221.0 &85.3$\%$ \\
\small TR-LSTM \cite{tensorR}  & 0.267M &221.8 &86.9$\%$ \\
\small HT-LSTM \cite{yin2020compressing}	&0.266M&	222.7 &	\textbf{87.9$\%$}\\
\hline
\small VIB-LSTM(ours) & \textbf{0.1778M} & \textbf{332.2}  &85.4$\%$ \\
\hline

\end{tabular}

\end{center}

\caption{Performance comparison between various LSTM compression methods applied to end-to-end LSTM architecture with the UCF11 dataset. VIB-LSTM outperforms other compressed methods as it prunes away a significantly larger number of redundant parameters while maintaining comparable validation accuracy.}
\label{tab0}

\end{table}

\section{Results and Observations}
We evaluate our end-to-end VIB-LSTM architecture on UCF11 and CNN-VIB-LSTM architecture on all three datasets. Many tensor decomposition based LSTM compression methods \cite{tensorR} compress only the input-to-hidden transformation matrix and compare only the number of corresponding matrix's parameters. We instead compare the total number of LSTM parameters. To compare the extent of compression, we use the compression ratio, which is the ratio of the total number of LSTM parameters in the uncompressed model to that in the compressed model. 

\begin{table}[t]
\begin{center}
\begin{tabular}{p{2.5cm}|p{1.5cm}|p{1.4cm}|p{1.1cm}}
\hline
\small
\textbf{\small Methods}&\textbf{\small Total}&\textbf{\small LSTM}&\textbf{\small Accuracy}\\
&\textbf{\small Parameters}&\textbf{\small Parameters}\\
\cline{1-4}
\small Two-Stream \cite{twolstm} &\small 134.3M &\small 5.9M &\small 94.6$\%$  \\
\small Attention-ConvLSTM\cite{ge2019attention} &\small 10.03M &\small 4.72M&\small 93.48$\%$ \\
\hline
\multicolumn{4}{c}{\small Comparison of LSTM compression methods}\\
\hline
\small Naive-LSTM  &\small 55.38M &\small 33.57M &\small 98.53$\%$  \\
\small TR-LSTM \cite{tensorR}  &\small 23.15M &\small 1.340M &\small 93.8$\%$ \\
\small ISS \cite{wen2018learning}&\small 21.97M &\small 0.16M &\small 94.6$\%$ \\
\small VIB-LSTM (ours) &\small 21.86M &\small \textbf{2052}  &\small \textbf{98.53$\%$} \\
\small VIB-LSTM+ISS &\small 21.86M &\small \textbf{1680} &\small  90.2$\%$ \\
\small EfficientNet +VIB-LSTM (ours) &\small 7.8M &\small \textbf{1680}&\small 96.6$\%$ \\
\hline
\end{tabular}
\end{center}
\caption{Comparison of parameters and accuracy of different LSTM compression methods and state-of art architectures on UCF11 dataset shows VIB-LSTM's amazing performance w.r.t both the compression and the validation accuracy.\label{tab2}}
\end{table}

\subsection{Comparison on UCF11 dataset}

On training end-to-end VIB-LSTM with UCF11, we obtain sizable reduction in the total number of LSTM parameters with validation accuracy comparable to the other LSTM compression methods, as shown in Table \ref{tab0}. Our experiments show that end-to-end VIB-LSTM uses only 1312 out of 57,600 values in the raw input tensor, for prediction. 

\begin{figure}
\includegraphics[width=8cm]{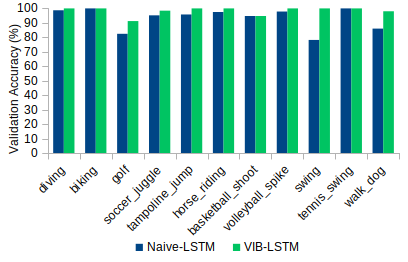}
   \caption{Performance comparison on the various classes of UCF11 dataset, between the baseline Naive-LSTM model \cite{tensorR} and our VIB-LSTM trained model.} 
\label{fig:compucf}
\end{figure}

\begin{figure}[t]
\begin{center}
\includegraphics[width=8cm]{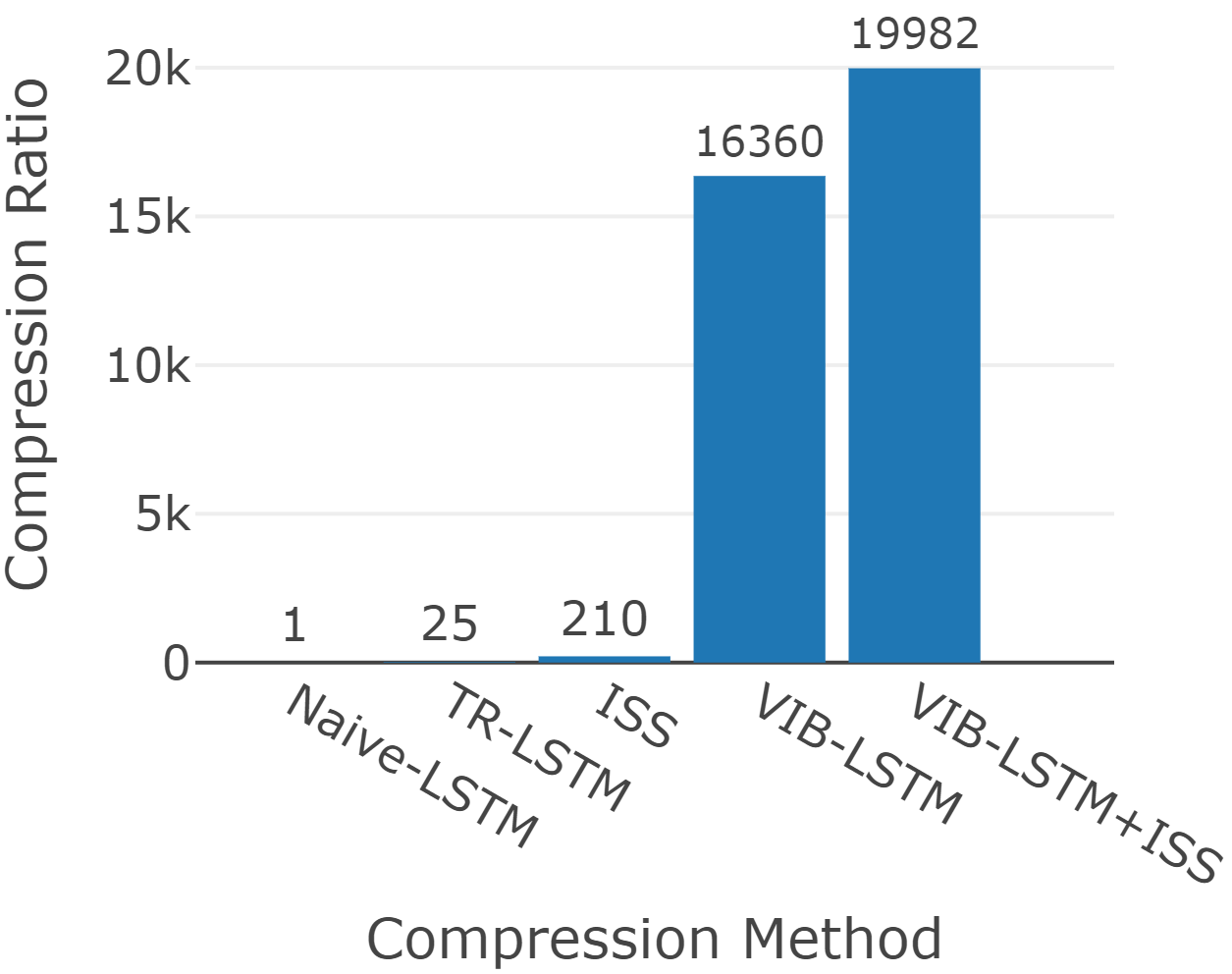}
\end{center}
   \caption{Comparison of LSTM Compression ratios achieved by different LSTM compression methods on same CNN-LSTM architecture with UCF11. Our approach surpasses other methods by a sizeable difference.}
\label{bar}
\end{figure}

\begin{table*}[ht]
\begin{center}
\begin{tabular}{p{5cm}|l | l|l|l | l|l}
\hline
\textbf{Methods} & \textbf{Total}  & \textbf{LSTM} & \textbf{Accuracy} &  \textbf{Total}  & \textbf{LSTM} & \textbf{Accuracy} \\
& \textbf{Parameters} & \textbf{Parameters} & & \textbf{Parameters} & \textbf{Parameters} &\\
\hline
\multicolumn{1}{c|}{} &\multicolumn{3}{c|}{HMDB Dataset}&\multicolumn{3}{c}{UCF101 Dataset} \\
\hline
\small Attention-ConvLSTM \cite{ge2019attention} & \small 10.03M&\small 4.72M &\small 67.1$\%$ & \small 10.03M& \small 4.72M &\small 92.88$\%$ \\
\small Two-StreamI3D- &&&&&&\\ only imagenet-pretrained \cite{carreira2017quo} &\small 25M & - &\small 66.4$\%$  &\small 25M & - &\small 93.4$\%$\\
\small TS-LSTM \cite{ma2019ts} &\small 53.5M &\small 9M &\small 69.0$\%$&\small 53.5M &\small 9.44M &\small \textbf{94.3}$\%$ \\
\hline
\multicolumn{7}{c}{Comparison of LSTM compression methods}\\
\hline
\small Naive-LSTM (uncompressed)&\small 55.38M &\small 33.57M &\small 62.9$\%$ &\small 48.41M &\small 9.44M &\small 92.6$\%$\\
\small Naive-TS-LSTM (uncompressed)&\small 55.38M &\small 33.57M &\small 69.0$\%$ &\small 48.41M &\small 9.44M &\small 93.5$\%$\\
\small TR-LSTM \cite{tensorR}&\small 55.38M &\small 1.340M &\small 63.8$\%$&-&-&- \\
\small ISS \cite{wen2018learning}& - & - & - &\small 59.57M &\small 1.7M &\small 90.2$\%$ \\
\small VIB-LSTM (Ours)&\small 21.86M &\small \textbf{0.41M} &\small \textbf{64.5$\%$} &\small 21.86M &\small \textbf{0.46M} & \small 92.2$\%$ \\
\small TS-VIB-LSTM (Ours)&\small 21.86M &\small \textbf{0.41M} &\small 68.16$\%$&\small 21.86M &\small \textbf{0.46M} &\small 93.15$\%$ \\
\hline
\end{tabular}s
\end{center}
\caption{Performance comparison of different methods on split 1 of HMDB51 and UCF101 datasets. VIB-LSTM compresses Naive-LSTM by $81 \times$ for HMDB51 and by $20 \times$ for UCF101.\label{tab3}}
\end{table*}

For CNN-LSTM based architecture, we use Inception\_v3 as the feature extractor \cite{szegedy2016incepv3} and 2048 dimensional hidden state vector for the LSTM, with similar settings as in \cite{tensorR}. By incorporating VIB-LSTM into the Naive CNN-LSTM model, we achieve $ 16,360 \times$ reduction in LSTM size with significant improvement in accuracy, as shown in Table \ref{tab2}. Thus, the proposed VIB-LSTM model reduces overfitting and classifies several actions better than the Naive-LSTM model, as shown in Figure \ref{fig:compucf}.
Compared to \cite{tensorR} and \cite{wen2018learning}, VIB-LSTM based models achieve higher compression ratio with accuracy comparable to that of the Naive-LSTM model, as shown in Figure \ref{bar}. After training with VIB-LSTM, the input and the hidden state vector dimensions of LSTM come down from 2048 to 49 and 9, respectively.
Moreover, combining the group lasso method \cite{ye2018learning} with VIB-LSTM leads to even more significant compression with a little drop in the accuracy (refer to Table \ref{tab2}).
Moreover, several FC parameters following the redundant hidden states of LSTM also become redundant, thus further reducing the total number of model parameters.
We also perform experiments by taking EfficientNet-b1 \cite{tan2019efficientnet} extracted features with VIB-LSTM, which yields even smaller model which is much suitable for deployment on small edge devices.

\subsection{Comparison on HMDB51 dataset}
\begin{figure}[t]
\begin{center}
\includegraphics[width=8cm]{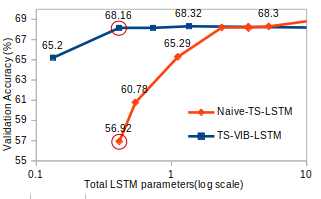}
  \caption{Performance comparison of Naive-TS-LSTM and TS-VIB-LSTM for comparable number of parameters validated on HMDB51 dataset. VIB model performs much better for the same 0.41M total LSTM parameters marked by the circle.} 
\label{fig:naivevsVIB}
\end{center}
\end{figure}

In the experiments with HMDB51, we randomly sample 25 frames from each video and save the Inception\_v3 extracted features to train the model. A random sampling at every epoch reduces overfitting as the model sees more variation in the data and generalizes better. 
In Table \ref{tab3}, we compare the CNN-VIB-LSTM model with the baseline uncompressed Naive-CNN-LSTM model and other compression methods. Our VIB-LSTM achieves better accuracy with a compression ratio of $81\times$ compared to $25\times$ in the TR-LSTM.
Temporal segment(TS) batch norm \cite{ma2019ts}, when used in conjunction with Naive-LSTM, reduces overfitting, and improves the validation accuracy. 
We further compare our VIB-LSTM trained model with state-of-art architectures. Two-Stream I3D \cite{carreira2017quo} comprising 25M parameters of inflated Inception\_v1 gives 66.4$\%$ accuracy. Our TS-VIB-LSTM model gives much higher accuracy with about 3.14M lesser number of parameters and with much lesser flops due to 2D CNN in our model rather than 3D CNN. TS-VIB-LSTM has 11 times lower LSTM parameters than Attention-based convolutional-LSTM \cite{ge2019attention}. 
The LSTM hidden state vector size is a hyperparameter, challenging to tune while working on large datasets. Our proposed method eliminates this problem. After training with a reasonably significant LSTM hidden state size, VIB-LSTM can bring down the hidden state vector to a size relevant to prediction. For a comparable number of parameters, VIB-LSTM-compressed models perform much better than naive models, as shown in Figure \ref{fig:naivevsVIB}.

\subsection{Comparison on UCF101 dataset}
\begin{figure}[t]
\begin{center}
\includegraphics[width=8cm]{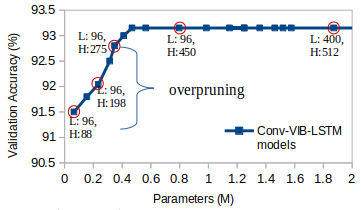}
  \caption{Variation in performance of the model with respect to the degree of compression obtained for different values of $\beta$ with UCF101. L and H denote the effective size of the VIB-LSTM input vector and VIB-LSTM hidden state vector, respectively.} 
\label{fig:overp}
\end{center}
\end{figure}

As with HMDB51 setup, both spatial RGB frame features and temporal optical flow features of UCF101 videos extracted using Inception\_v3 are concatenated before passing on to the LSTM models. The baseline models, Naive-LSTM and Naive-TS-LSTM, have 512 hidden states each. 
The VIB-LSTM compressed models have negligible accuracy degradation compared to the uncompressed Naive-LSTM models, while more than 20 times a reduction in the number of LSTM parameters is achieved. 
Using only 3.8$\%$ of the original input features, VIB-LSTM has 270 prediction-relevant hidden states. We make use of the temporal segment (TS) batch norm \cite{ma2019ts}, which improves the accuracy of TS-VIB-LSTM over VIB-LSTM, as shown in Table \ref{tab3}. 
 We experiment with different hyperparameter values $\beta$ (usually, we take $\beta > 1$). The amount of compression is proportional to the value of beta. We observe that accuracy degrades significantly after a particular point with increasing compression and decreasing the number of LSTM parameters due to large values of $\beta$ ($ \geq 2$), as shown in Figure \ref{fig:overp}. This behavior is similar to the other two datasets as well.
Further, our TS-VIB-LSTM model achieves comparable accuracy with much lesser total parameters than Two-Stream I3D \cite{carreira2017quo} and Attention-based convolutional-LSTM \cite{ge2019attention}. 

\subsection{Ablation Experiments}
For the CNN-LSTM architecture taken in our experiments, we perform ablation experiments using UCF11 to analyze the particular importance of VIB-LSTM and feature masking VIB layer. We first use VIB-LSTM only, without the feature mask layer. This allows a reduction of the hidden state size with VIB-LSTM but constricts a reduction in the LSTM input size. We observe that VIB-LSTM without compressed input features can efficiently prune many redundant parameters, as it results in a compression ratio of about 510 with accuracy over 98.5$\%$. Next, we use only the feature mask layer with Naive-LSTM and observe that it barely yields a compression ratio of 2. The ablation results (Refer Table \ref{table:ablation}) show that VIB-LSTM alone can induce huge sparsity in the LSTM parameter matrix; however, we get much better performance on using VIB-LSTM clubbed with the feature mask layer. 

\subsection{Inference on Raspberry Pi}
Raspberry Pi model 3 with 1GB RAM, 64-bit quad-core Arm Cortex-A53 CPU is selected to test the inference speed-up using our approach over Naive-LSTM for the same task. 
We test the Naive-LSTM and VIB-LSTM model using a single video and a batch of 10 videos from UCF11, running on only one processor core to avoid automated parallelism. Features extracted from Inception\_v3 CNN are given input to both Naive-LSTM(33.5 Million parameters) and VIB-LSTM(2052 parameters) separately. VIB-LSTM shows up to \textbf{100x} speed up over Naive-LSTM.

\begin{table}[t]
\begin{center}
\begin{tabular}{p{2.5cm}|p{1.5cm}|p{1.4cm}|p{1.1cm}}
\hline
\small

\textbf{\small Network \quad\quad\quad Architecture}& \textbf{\small Total Params}&\textbf{\small LSTM Params}&\textbf{\small Accuracy}\\
\cline{1-4}
\small Naive-LSTM  &\small 55.38M &\small 33.57M &\small 98.53$\%$  \\
\small w/o VIB on LSTM &\small 38.81M &\small 17.06M  &\small 98.53$\%$ \\
\small w/o feature mask layer &\small 21.92M &\small 65856  &\small 98.53$\%$ \\
\small VIB-LSTM with feature mask layer &\small 21.86M &\small \textbf{2052}  &\small \textbf{98.53$\%$} \\
\hline
\end{tabular}
\end{center}
\caption{Ablation experiment results obtained by first removing feature mask layer and then by removing VIB-LSTM.}
\label{table:ablation}
\end{table}

\section{Conclusion}
In this paper, we present a generic RNN compression technique based on the VIB theory. Specifically, we propose a compression scheme that extracts the prediction-relevant information from the input features. To this end, we formulate a loss function aimed at LSTM compression with end-to-end and CNN-LSTM architectures for HAR. We minimize the formulated loss function to show that our approach significantly compresses the baseline's over-parameterized LSTM structural matrices. Our approach improves the performance of large-capacity baseline models by reducing the problem of overfitting in them, and reduces both the memory footprint and the computational requirements. Thus, our approach can produce models suitable for deployment on edge devices, which we show by deploying our VIB-LSTM trained model on Raspberry Pi and inferencing it.\\
Further, we show that our approach can be effectively used with other compression methods to obtain even more significant compression with a little drop in accuracy. More study is required on our assumption of uncorrelated features and hidden states. Our approach can be extended to several other sequence modeling applications, such as machine translation and speech synthesis.

{\small
\bibliographystyle{ieee_fullname}
\bibliography{egbib}
}

\end{document}